%% file: main.tex
\documentclass[sigconf]{acmart} 
\usepackage{graphicx}
\usepackage{subcaption}
\usepackage{xspace}
\usepackage{multirow}
\usepackage{enumitem}
 
\usepackage[table]{xcolor}

\usepackage[compact]{titlesec}

\titlespacing*{\section}{0pt}{0.8ex}{0.2ex}
\titlespacing*{\subsection}{0pt}{0.3ex}{0.1ex}
\usepackage{colortbl,array}
\usepackage{float}
\usepackage{needspace}

\usepackage{booktabs, colortbl, xcolor, array, geometry, graphicx}
\newcolumntype{G}{>{\columncolor{gray!30}}c} 
\newcolumntype{g}{>{\columncolor{gray!15}}c} 

\AtBeginDocument{%
  }

\newcommand{\data}{\textsc{Real-E}\xspace}

\newcommand{\metricf}{$C_f$\xspace}
\newcommand{\link}{https://github.com/YueW26/Real-E}
\newcommand{\datalink}{https://zenodo.org/records/15685930}

\acmDOI{https://doi.org/10.1145/3746252.3761637}  

\copyrightyear{2025}
\acmYear{2025}
\setcopyright{acmlicensed}
\acmConference[CIKM '25] {Proceedings of the 34th ACM International Conference on Information and Knowledge Management}{ November 10--14, 2025}{Seoul, Republic of Korea.}
\acmBooktitle{Proceedings of the 34th ACM International Conference on Information and Knowledge Management (CIKM '25), November 10--14, 2025, Seoul, Republic of Korea}
\acmISBN{979-8-4007-2040-6/2025/11}

\acmSubmissionID{crp6954}
\begin{document}

\title{\data: A Foundation Benchmark for Advancing Robust and Generalizable Electricity Forecasting}

\author{Chen Shao}
\authornote{Both authors contributed equally to this research.} %
\orcid{0000-0002-7807-1090}
\affiliation{%
  \institution{Karlsruhe Institute of Technology}
  \city{Karlsruhe}
  \country{Germany}}
\email{chen.shao2@kit.edu}

\author{Yue Wang}
\authornotemark[1]
\orcid{0009-0001-8566-3545}
\affiliation{%
  \institution{Karlsruhe Institute of Technology}
  \city{Karlsruhe}
  \country{Germany}}
\email{pg9697@partner.kit.edu}

\author{Zhenyi Zhu}
\orcid{0009-0000-4625-7088}
\affiliation{%
  \institution{The Hong Kong University of Science and Technology}
  \city{Hong Kong}
  \country{China}}
\email{zzhubh@connect.ust.hk}

\author{Zhanbo Huang}
\orcid{0009-0009-6042-0174}
\affiliation{%
 \institution{Karlsruhe Institute of Technology}
 \city{Karlsruhe}
 \country{Germany}}
\email{zhanbo.huang@rwth-aachen.de}

\author{Sebastian Pütz}
\orcid{0009-0009-8468-4166}
\affiliation{%
  \institution{Karlsruhe Institute of Technology}
  \city{Karlsruhe}
  \country{Germany}}
\email{sebastian.puetz@kit.edu}

\author{Benjamin Schäfer}
\orcid{0000-0003-1607-9748}
\affiliation{%
  \institution{Karlsruhe Institute of Technology}
  \city{Karlsruhe}
  \country{Germany}}
\email{benjamin.schaefer@kit.edu}

\author{Tobias Käfer}
\orcid{0000-0003-0576-7457}
\affiliation{%
  \institution{Karlsruhe Institute of Technology}
  \city{Karlsruhe}
  \country{Germany}}
\email{tobias.kaefer@kit.edu}

\author{Michael Färber}
\orcid{0000-0001-5458-8645}
\affiliation{%
  \institution{Karlsruhe Institute of Technology}
  \city{Karlsruhe}
  \country{Germany}}
\email{michael.faerber@tu-dresden.de}

\renewcommand{\shortauthors}{Chen et al.}

\begin{abstract}
Energy forecasting is vital for grid reliability and operational efficiency. Although recent advances in time series forecasting have led to progress, existing benchmarks remain limited in spatial and temporal scope and lack multi-energy features. This raises concerns about their reliability and applicability in real-world deployment. To address this, we present the \data dataset, covering over 74 power stations across 30+ European countries over a 10-year span with rich metadata. Using \data, we conduct an extensive data analysis and benchmark over 20 baselines across various model types. We introduce a new metric to quantify shifts in correlation structures and show that existing methods struggle on our dataset, which exhibits more complex and non-stationary correlation dynamics. Our findings highlight key limitations of current methods and offer a strong empirical basis for building more robust forecasting models.


\end{abstract}


\maketitle

\section{Introduction}

\begin{figure*}[t]
    \centering
    \includegraphics[width=\linewidth, trim=0 150 0 0, clip]{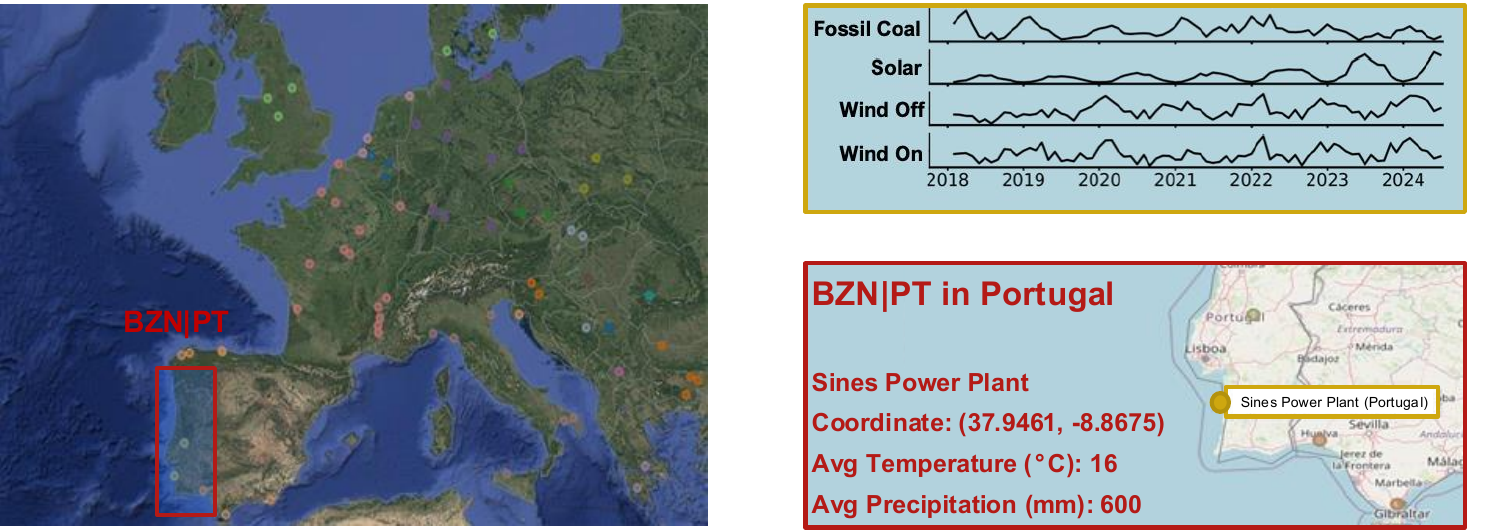}
    \captionsetup{skip=2pt} 
    \caption{\data is the largest electricity dataset to date, covering 10.18 million $\text{km}^2$ across 39 countries (colored dots) and spanning over 20 generation types (yellow boxes). Each power plant is accompanied by detailed metadata (red boxes). It offers a rigorous foundation for advancing multivariate electricity forecasting research.}
    \label{fig:cover-image}
\end{figure*}

Energy forecasting plays a critical role in modern power systems, supporting key tasks such as grid stability, energy planning and renewable energy integration. Recent advances in multivariate time forecasting (MTSF) have shown strong potential in modeling the complex spatiotemporal dependencies. Recurrent Neural Network (RNN)~\cite{lai2018modeling}, Convolutional Neural Network (CNN)~\cite{bai2018empirical} and Transformers~\cite{kitaev2020reformer} primarily incorporate complex spatial and temporal dependencies using either an auto-regressive model or a global attention mechanism. Graph Neural Network (GNN)-based approaches incorporate spatial correlations by modeling spatial dynamics from inter-series correlations~\cite{wu2019graph, wu2020connecting, liu2022tpg}. 

Despite the improved performance, these methods have been predominantly evaluated on small-scale datasets, such as electricity \cite{electricityDataset} and solar \cite{lai2018modeling}. They focus solely on a \emph{single generation type from a single country with limited temporal coverage}. However, modern power systems rely on an integrated mix of energy sources including wind, hydro, nuclear and coal that interact with one another. These interactions are further influenced by policy constraints and seasonal dynamics. Between such time-evolving correlation, each time series also exhibits seasonal patterns shaped by weather conditions, temperature fluctuations and geographic diversity. However, these complex data factors and their impact on the performance of current forecasting models have not yet been well studied.

Our work addresses these issues by first introducing \textbf{\data}, the largest electricity dataset to date, to support the development of robust and efficient methods for forecasting (Fig.~\ref{fig:cover-image}). We conduct a thorough data analysis to compare the proposed approach with existing datasets and then perform extensive benchmark experiments. Based on the empirical results, we summarize key research challenges inherent in current approaches. Our main contributions include:

\begin{enumerate}[leftmargin=*, itemsep=0pt, topsep=0pt]
    \item \textbf{Largest Real-world Electricity Dataset} The proposed \data dataset offers unparalleled spatiotemporal breadth, encompassing 20 distinct energy types across 39 European countries over a period of up to 10 years. This dataset provides a foundational resource for advancing this field to address the challenges of generalizability and efficiency in real-world deployment. 
    \item \textbf{Quantifying Temporal Correlation Evolution:} Based on our data analysis, we identify unique data characteristics of the proposed dataset: more complex time-varying complementary patterns among multiple energy sources. We introduce a metric to quantify this and compare it with existing datasets and demonstrate that our dataset exhibits significantly higher volatility and structural complexity in its correlation dynamics.
    
    \item \textbf{Comprehensive Benchmarking and Revealed Limitations.}
    We benchmarked over 20 approaches across diverse architectural families on \data. Our empirical findings reveal that state-of-the-art Transformer-based and Spectral GNN models struggle to generalize under conditions of pronounced correlation dynamics. We make our dataset and benchmark publicly available at \href{https://zenodo.org/records/15685930}{[Real-E Link]} and \href{https://github.com/YueW26/Real-E}{[Benchmark Link]}.
\end{enumerate}

\section{Proposed \data}

We build \data from the European Network of Transmission System Operators for Electricity (ENTSO-E) Transparency Platform~\cite{entsoe} updated to June 2024. We excluded time series with substantial missing values to maximize the coverage of generation types and temporal alignment across energy types. Please refer to our repository for more details. %

We compare \data with existing benchmarks including ETT \cite{zhou2020informer}, Electricity~\cite{electricityDataset} and Solar-Energy~\cite{lai2018modeling} from three perspectives: \textbf{(1) Realistic Setting:} \data reflects multiple challenges in energy systems on a real-world scale. It covers the full electricity lifecycle (see Tab. \ref{tab:data-compare}), including \emph{Generation} (Energy Production and production forecasts), \emph{Transmission} (Power Transfer over borders between areas),  \emph{Balancing} (Regulation Energy to ensure the electrical transmission grid is in balance), \emph{Market} (Trade and Price) and \emph{Load} (Data about power consumption). The data spans multiple energy sources such as wind, solar, hydro, thermal, nuclear and pumped storage. \textbf{(2) Extensive Temporal and Geographic Coverage:} As detailed in Tab.~\ref{tab:data-compare}, \data spans a full decade from \emph{2014 to 2024} of multiple temporal resolutions ranging from 15-min to hourly records. In addition, \data provides broad geographic representation across 39 European countries, including energy-specific correlation structures (Sec~\ref{sec:data_analysis}). \textbf{(3) Rich Operational Metadata:} Each time series is accompanied by contextual information. For example, the dataset includes spatial descriptors (e.g., coordinates or bidding zones) that enable geographical clustering, as well as system-level attributes (e.g., transmission distance, voltage level, or grid topology) that provide insights into the operational context of the measurements (see Fig.~\ref{fig:cover-image}).
\begin{table}[h]
\centering
\tiny
\renewcommand{\arraystretch}{0.85}
\setlength{\abovecaptionskip}{-0.5em}  
\setlength{\belowcaptionskip}{-1em}    
\caption{Comparing with existing datasets, \data offers longer duration (Dur.), finer resolution (Res.), longer sequences (Len.), more energy categories (EC) and broader country coverage (C).}
\label{tab:data-compare}
\resizebox{0.48\textwidth}{!}{%
\begin{tabular}{@{}c@{\hskip 4pt}l@{\hskip 4pt}c@{\hskip 3pt}c@{\hskip 3pt}c@{\hskip 3pt}c@{\hskip 3pt}c@{}}
\toprule
& \textbf{Name} & \textbf{Dur.} & \textbf{Res.} & \textbf{Len.} & \textbf{EC} & \textbf{C} \\
\midrule
\multirow{4}{*}{\rotatebox[origin=c]{0}{\parbox{1.5cm}{\centering\fontsize{6}{6}\selectfont\textcolor{gray}{\textbf{Existing\\Benchmark}}}}}
& \textcolor{gray}{Solar}       & \textcolor{gray}{1 y}    & \textcolor{gray}{10 min}  & \textcolor{gray}{52,560}  & \textcolor{gray}{1}  & \textcolor{gray}{1} \\
& \textcolor{gray}{Electricity} & \textcolor{gray}{3 y}    & \textcolor{gray}{1 hour}  & \textcolor{gray}{26,304}  & \textcolor{gray}{1}  & \textcolor{gray}{1} \\
& \textcolor{gray}{ETTh1/ETTh2}       & \textcolor{gray}{20 mon} & \textcolor{gray}{1 hour}  & \textcolor{gray}{14,400}  & \textcolor{gray}{1}  & \textcolor{gray}{1} \\
& \textcolor{gray}{ETTm1/ETTm2}       & \textcolor{gray}{20 mon} & \textcolor{gray}{15 min}  & \textcolor{gray}{57,600}  & \textcolor{gray}{1}  & \textcolor{gray}{1} \\
\midrule
\multicolumn{7}{c}{\textbf{\data}} \\
\midrule
\multirow{4}{*}{\rotatebox[origin=c]{0}{Generation}} 
  & \textcolor{teal}{Actual-ByType}      & \textcolor{teal}{9.5 y} & \textcolor{teal}{15 min}  & \textcolor{teal}{>330k}   & \textcolor{teal}{20} & \textcolor{teal}{39} \\
  & Actual-ByUnit                        & 9.5 y & 1 hour  & ~8.7k   & 20 & 39 \\
  & Renewables-Forecast                  & 9.5 y & 15 min  & >330k   & 3  & 39 \\
  & Capacity-Annual                      & 9.5 y & 1 year  & ~10     & 20 & 39 \\
\midrule
\multirow{2}{*}{\rotatebox[origin=c]{0}{Load}} 
  & Actual                               & 9.5 y & 15 min  & >330k   & 20  & 39 \\
  & Forecast-WeekAhead                   & 9.5 y & 1 day   & ~3.4k   & 20  & 39 \\
\midrule
\multirow{2}{*}{\rotatebox[origin=c]{0}{Market}} 
  & Price-QuarterHourly                  & 9.5 y & 15 min  & >330k   & 20 & 39 \\
  & Price-Hourly                         & 9.5 y & 1 hour  & ~8.7k   & 20 & 39 \\
\midrule
\multirow{2}{*}{\rotatebox[origin=c]{0}{Transmission}} 
  & Capacity-Forecast                    & 9.5 y & 1 hour  & ~8.7k   & 20 & 39 \\
  & Flow-Actual                          & 9.5 y & 1 hour  & ~8.7k   & 20 & 39 \\
\midrule
\multirow{2}{*}{\rotatebox[origin=c]{0}{Balancing}} 
  & Energy-Activated                     & 9.5 y & 15 min  & >330k   & 20 & 39 \\
  & System-Imbalance                     & 9.5 y & 1 hour  & ~8.7k   & -- & 39 \\
\bottomrule
\end{tabular}%
}
\end{table}

\section{Data Analysis}
\label{sec:data_analysis}

We reveal the time-varying properties of energy forecasting across temporal and cross-energy dimensions. For data analysis, we use the dataset recording electricity production from different energy sources in Germany and France(\textcolor{teal}{highlighted in Tab. \ref{tab:data-compare}}).

\textbf{Seasonal Complementary Patterns.} Figure~\ref{fig:Seasonal} illustrates the annual trends of two representative energy sources, brown coal and solar power in Germany. These sources exhibit a clear seasonal complementarity: brown coal reaches its peak during the autumn and winter months, while solar power peaks in summer. Germany’s overall electricity demand throughout the year displays a distinct bimodal pattern, with each peak predominantly supported by a different energy source. Such short and long-term time-evolving correlation continually challenges the prevailing short-window time series models, exposing their limitations in capturing long-range and evolving dynamics.

\textbf{Cross Energy Time-Evolving Correlation.} We generalize the analysis to encompass \emph{all major energy sources} in Germany. Figure~\ref{fig:Correlation} reveals dynamic shifts in the inter-dependencies between multiple energy resources. These evolving correlation structures reflect fundamental changes in the internal coordination of the energy system. As the figure highlights, we observe a clear intensification of correlations in the upper-left block informed by biomass, brown coal and gas. 

\textbf{Quantify Time-Evolving Correlation} To characterize such a time-evolving correlation pattern, we propose two metrics as follows: \\
\emph{Temporal Graph Volatility (TGV)} quantifies the structural variation in the adjacency matrix $A$ between two adjacent time steps $t$ via the Frobenius norm. A higher value indicates more frequent structural transitions.
{\setlength{\abovedisplayskip}{1pt}   
 \setlength{\belowdisplayskip}{1pt}   
 \begin{equation}
        TGV = \sum_{t=0}^{T-1} \|A_{t+1} - A_t\|_F = \sum_{t=1}^{T} \sqrt{\sum_{i=1}^{m} \sum_{j=1}^{n} |a^{t+1}_{ij} - a^{t}_{ij}|^2}.
\label{eq:drift}
\end{equation}}
\noindent \emph{Graph Spectral Divergence (GSD) \metricf} indicates periods of structural volatility via spectral distances. It identifies correlation regime shifts, such as policy events or grid reconfigurations. We compute the Laplacian spectral distances of matrices as follows: 
\setlength{\parskip}{-0.2em}
{\setlength{\abovedisplayskip}{1pt}   
 \setlength{\belowdisplayskip}{1pt}   
\begin{equation}
GSD = \left\| \lambda(L_t) - \lambda(L_{t-1}) \right\|_2, L_t = D_t - A_t
\label{eq:shift}
\end{equation}}
where $D_t = \text{diag}(A_t \mathbf{1}), A_t$ are the degree and adjacent matrix. The spectral topology shift is measured as the Euclidean distance between the sorted Laplacian eigenvalues. 

\noindent \textbf{Metrics in Practice} We report average metric values across all time steps on existing benchmarks. For the existing datasets, Temporal Graph Volatility (TGV) achieves an average of 0.5822, approaching a maximum of 0.6245, while Graph Spectral Divergence (GSD) averages 1.6298 and peaks at 1.8027. In contrast, our dataset attains higher values, with TGV averaging 1.03335 and peaking at 1.0969, and GSD averaging 2.3237 and peaking at 2.7289. For empirical analysis, we focus on Germany and France as representative cases.

\begingroup
\setlength{\textfloatsep}{6pt} 
\begin{figure}[t!]
\begin{subfigure}[b]{0.44\linewidth}
    \centering
    \includegraphics[height=\linewidth, width=\linewidth]{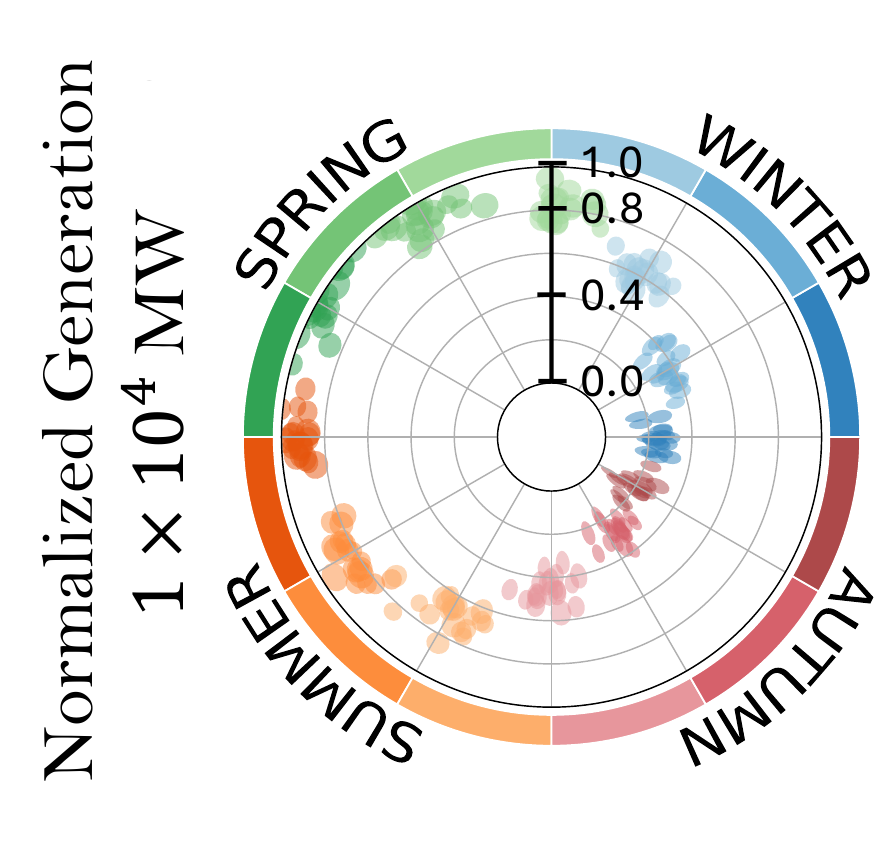}
    \captionsetup{skip=1.2pt} 
    \caption{}
    \label{fig:SeasonalCircular}
\end{subfigure}
\hfill
\begin{subfigure}[b]{0.55\linewidth}
    \centering
    \includegraphics[height=0.72\linewidth, width=\linewidth]{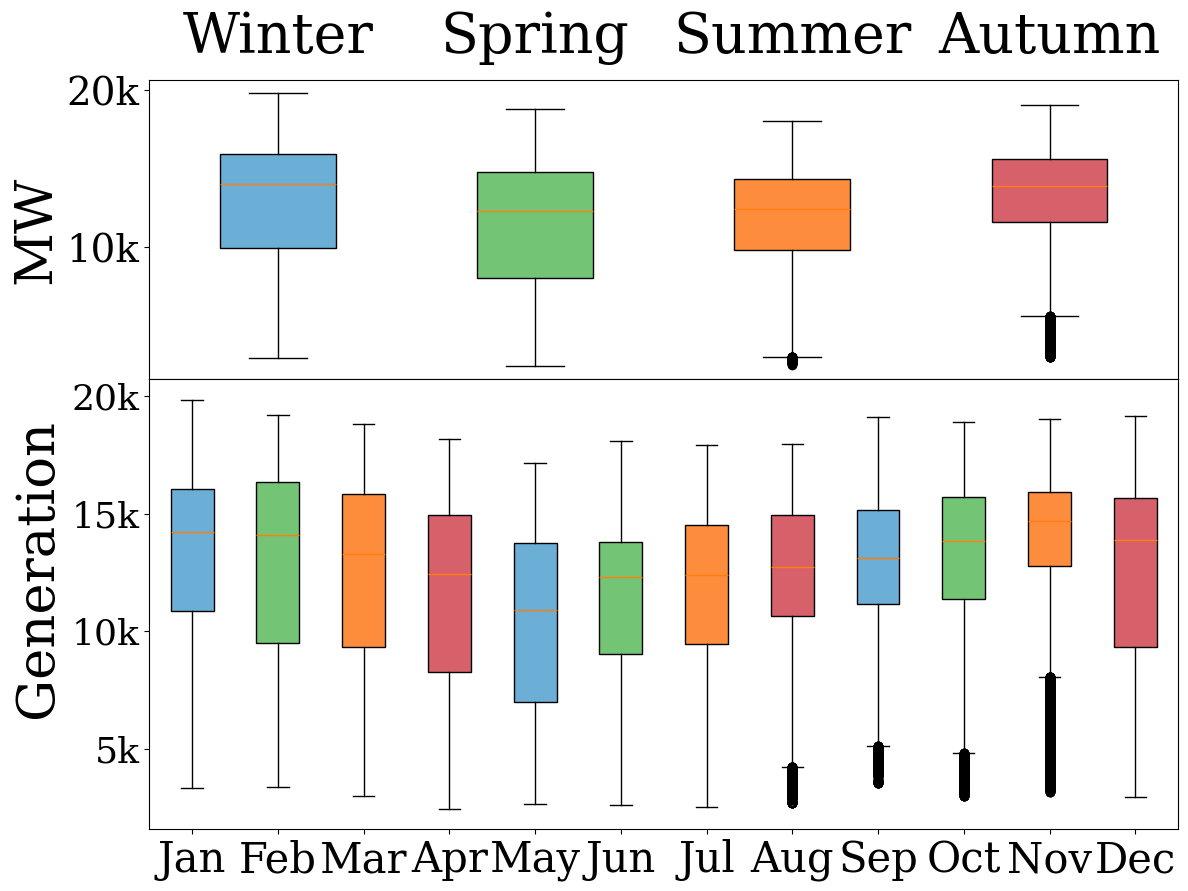}
    \captionsetup{skip=1pt} 
    \caption{}
    \label{fig:MonthlyBox}
\end{subfigure}
\captionsetup{skip=2.5pt} 
\caption{\textit{We use a polar plot (a) and a boxplot (b) to illustrate seasonal generation patterns of two complementary energy sources in Germany: solar and brown coal. (a) Solar generation shows a clear seasonal trend, with higher output in spring and summer, peaking in June and reaching its minimum in December — primarily driven by solar irradiance. (b) In contrast, brown coal generation peaks in autumn and winter and drops in warmer months, reflecting increased demand during colder periods.}}
\label{fig:Seasonal}
\end{figure}
\begin{figure}[h]
\centering
\includegraphics[width=\linewidth, trim=0 5 0 0, clip]{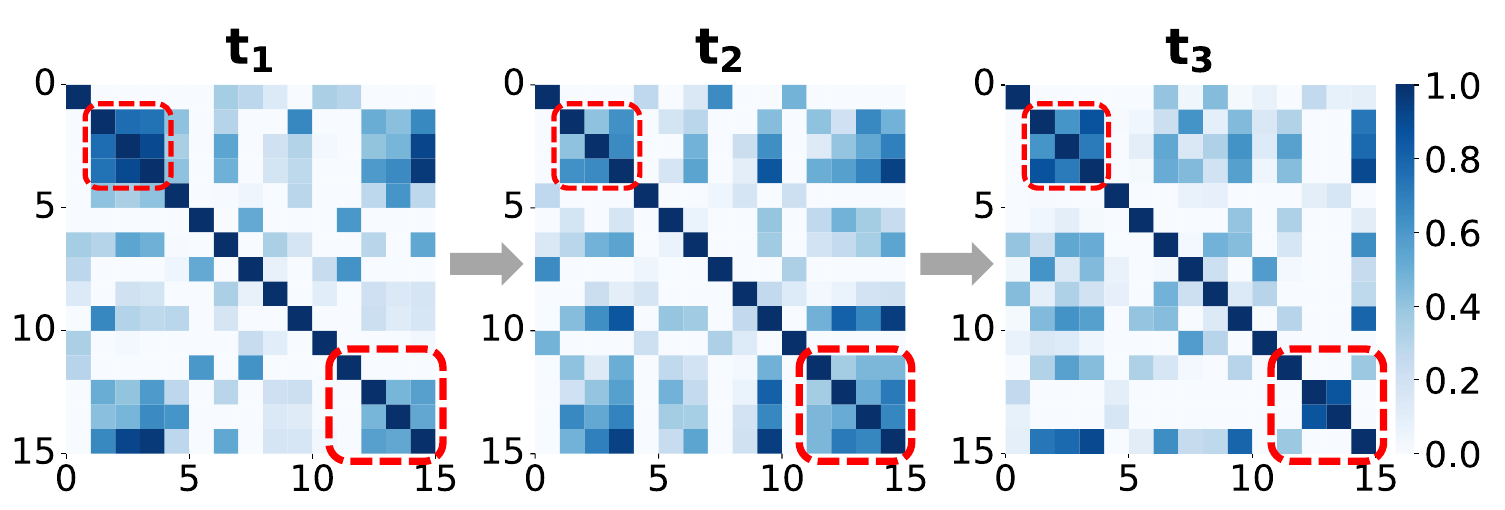}
\captionsetup{skip=1pt}
\
\caption{We present a series of Pearson correlation matrices across consecutive time windows. The bottom-right block highlights the strong, mutually inclusive correlations among lignite, gas and hard coal generation.}
\label{fig:Correlation}
\end{figure}
\endgroup

\section{Experiments}
In this section, we compare extensive approaches on both \data and public benchmark, aiming to answer the following research questions: \textbf{RQ1:} How well do existing forecasting models generalize to the more complex, real-world structure of \data? \textbf{RQ2:} Among different architectural paradigms, which models are most effective for forecasting under \data's dynamic spatial-temporal structure? 

\textbf{Dataset \& Metrics. } We choose a subset of proposed datasets, \data(\emph{Germany}, \emph{France}) and two public datasets (electricity and solar). It records electricity production in Germany and France in all generation types for all available years. We use the consistent data splits (70\%/20\%/10\%) for all models and report \emph{Mean Absolute Error (MAE)} and \emph{Root Mean Squared Error (RMSE)} of forecasting horizons with 12 time steps on five random seeds with enabled early stop. Due to space constraints, we provide the details of data preprocessing and splitting in  \href{Github}{Link}.

\textbf{Hyperparameter Tuning.} We performed a grid search over a wide range of hyperparameter values for all reported models. Due to limited space, we only report a subset of the tuned hyperparameters, which includes but not limited to: learning rate $10^{(-2\sim-4)}$, hidden embedding size $2^{8\sim10}$, batch size $2^{4\sim9}$, number of encoder and decoder layers $1\sim3$, number of attention heads $2\sim8$, kernel sizes $3\sim7$ and rolling window sizes $12\times(1\sim3)$. For the electricity, solar and ETT datasets, we report the best results obtained from \cite{yi2023fouriergnn, wu2023timesnet}.


\textbf{Baselines.} We evaluate 20 baselines reported in Tab. \ref{tab:results_mae_rmse}, which includes three classical statistical methods \cite{sims1980macroeconomics, box1970time, box2015time}, three MLP-based approaches \cite{zeng2023dlinear, oreshkin2020nbeats, wang2024timemixer} and several RNN/CNN-based architectures \cite{sen2019think, lai2018modeling, bai2018empirical, zhang2017stock}. We also include three advanced paradigms: 1)  Transformer-based models \cite{kitaev2020reformer, zhou2021informer, wu2021autoformer, zhou2022fedformer}, which optimize temporal modeling via optimized attention mechanisms; 2) Spectral GNN models \cite{cao2020spectral, Jin2023Towards, yi2023fouriergnn}, which leverage nonlinear filtering in the spectral domain to capture spatial correlations; 3) Spatial graph-based approaches \cite{wu2019graph, wu2020connecting, liu2022tpg}, which rely on localized message passing mechanisms to dynamically model and adapt to spatial dependencies.

\input{performance_tab}
\textbf{RQ1: Ranking Cross-Dataset Generalization} Table~\ref{tab:results_mae_rmse} reports the average MAE and RMSE for a forecasting horizon of 12 time steps of partial results. Other detailed results and information can be found at our  \href{\link}{Github}. We observed that a consistent trend emerges; many models fail to maintain performance across datasets with varying temporal dynamics and correlation complexity. For instance, the average MAE of the 4 Transformer-based models deteriorated from 0.117 to 0.215, reflecting an 85.4\% increase in error. Spectral GNN, a method that consistently outperforms other approaches in existing benchmarks (highlighted in the Electricity and Solar datasets), experiences a performance degradation of approximately 16.20\%.

\textbf{RQ2: Rank for the Performance}  On the proposed dataset, the top three performing models are predominantly from the Spatial GNN category. Its design includes dynamic graph modeling that demonstrates the greatest robustness, consistently delivering stronger generalization across diverse datasets. This superiority can likely be attributed to the rich spatial correlations and temporal patterns present in the data, which enable the graph-based neurons to be effectively trained. In contrast, this advantage is less evident on traditional datasets. While Transformer-based models and certain Spectral GNNs perform comparably on earlier benchmarks, they fall short against Spatial GNN architectures on the proposed dataset—even under carefully tuned parameter settings.

\textbf{Summary} Our findings empirically reveal the limitation of the current transformer-based forecasting model: global attention mechanism struggles to generalize in the presence of rapidly shifting dependencies, which is crucial in real-world model deployment, while Spatial GNNs consistently outperform other models by explicitly modeling correlations through graph structures and dynamically adapting to evolving spatial patterns.

\section{Conclusions}
We evaluated the generalization ability of advanced models on real-world electricity datasets with challenging structural complexity. To support reproducibility, we have released all datasets and code under the Creative Commons Attribution 4.0 License (CC-BY 4.0), fully compliant with open data standards. Our work exposes current methodological limitations and underscores the urgent need for designs to effectively capture time-evolving dependencies in electricity forecasting. We believe our dataset lay a solid foundation for addressing this challenge and offer a viable direction for future research. 
\section{GenAI Usage Disclosure}
In the paper writing, generative AI tools were used to assist with language polishing and LaTeX formatting. In algorithm development, we use AI to support debugging suggestions. The AI tools were not used to generate original scientific content, original implementation, or influence the interpretation of results. Any text generated or suggested by AI has been carefully reviewed and edited by the authors to ensure precision and clarity.
\bibliographystyle{plainnat}
\bibliography{output}

\appendix

\end{document}

%% file: performance_tab.tex
\begin{table}[h]
\centering
\tiny
\renewcommand{\arraystretch}{0.85}
\setlength{\abovecaptionskip}{0.5em}  
\setlength{\belowcaptionskip}{-1em}    
\caption{We report the average MAE and RMSE evaluated on \data (subset) and compare them against existing benchmark datasets (Electricity and Solar). The * results are obtained from \cite{yi2023fouriergnn, wu2023timesnet}. Best model per benchmark highlighted in gray. }
\label{tab:results_mae_rmse}
\resizebox{0.99\linewidth}{!}{%
\setlength{\tabcolsep}{1.7pt}
\renewcommand{\arraystretch}{1.1}
\begin{tabular}{l|cccccccc}
\toprule
\textbf{Model} 
& \multicolumn{2}{c}{\textcolor{darkgray}{\textbf{Electricity}}} 
& \multicolumn{2}{c}{\textcolor{darkgray}{\textbf{Solar}}} 
& \multicolumn{2}{c}{\textbf{Germany}} 
& \multicolumn{2}{c}{\textbf{France}} \\
& \multicolumn{2}{c}{\textcolor{darkgray}{\textbf{0.5391}}} 
& \multicolumn{2}{c}{\textcolor{darkgray}{\textbf{0.6245}}} 
& \multicolumn{2}{c}{\textbf{1.0969}} 
& \multicolumn{2}{c}{\textbf{0.9698}} \\
& \textcolor{darkgray}{MAE} & \textcolor{darkgray}{RMSE} 
& \textcolor{darkgray}{MAE} & \textcolor{darkgray}{RMSE} 
& MAE & RMSE 
& MAE & RMSE \\
\midrule
\multicolumn{9}{l}{\textbf{Spatial GNN}} \\
GWaveNet \cite{wu2019graph} & \textcolor{darkgray}{0.094} & \textcolor{darkgray}{0.140} & \textcolor{darkgray}{0.183} & \textcolor{darkgray}{0.238} & \cellcolor{gray!60}\textbf{0.013} & \cellcolor{gray!60}{\textbf{0.028}} & \cellcolor{gray!60}\textbf{0.012} & \cellcolor{gray!30}{\textbf{0.025}} \\
MTGNN \cite{wu2020connecting} & \textcolor{darkgray}{0.077} & \textcolor{darkgray}{0.113} & \textcolor{darkgray}{0.151} & \textcolor{darkgray}{0.207} & \cellcolor{gray!30}\textbf{0.016} & \cellcolor{gray!30}{\textbf{0.034}} & \cellcolor{gray!60}\textbf{0.012} & \cellcolor{gray!60}{\textbf{0.023}} \\
TPGNN* \cite{liu2022tpg} &  \cellcolor{gray!30}\textcolor{darkgray}{\textbf{0.055}} & \cellcolor{gray!30}\textcolor{darkgray}{\textbf{0.080}} & \textcolor{darkgray}{\cellcolor{gray!30}\textbf{0.123}} & \textcolor{darkgray}{0.214} & \cellcolor{gray!15}{\textbf{0.099}} & \cellcolor{gray!15}{\textbf{0.173}} & \cellcolor{gray!15}{\textbf{0.089}} & \cellcolor{gray!15}{\textbf{0.158}} \\
\midrule
\multicolumn{9}{l}{\textbf{Spectral GNN}} \\
FourierGNN \cite{yi2023fouriergnn}  & \cellcolor{gray!60}\textcolor{darkgray}{\textbf{0.051}} & \cellcolor{gray!60}\textcolor{darkgray}{\textbf{0.077}} & \cellcolor{gray!60}\textcolor{darkgray}{\textbf{0.120}} & \cellcolor{gray!60}\textcolor{darkgray}{\textbf{0.162}} & \cellcolor{gray!15}\textbf{0.110} & \cellcolor{gray!15}\textbf{0.186} & \cellcolor{gray!15}\textbf{0.096} & \cellcolor{gray!15}\textbf{0.164} \\
StemGNN* \cite{cao2020spectral}  & \textcolor{darkgray}{0.070} & \textcolor{darkgray}{0.101} & \textcolor{darkgray}{0.176} & \textcolor{darkgray}{0.222} & 0.179 & 0.285 & \cellcolor{gray!15}\textbf{0.148} & 0.206 \\
TGGC \cite{Jin2023Towards}  & \textcolor{darkgray}{0.086} & \textcolor{darkgray}{0.127} & \textcolor{darkgray}{0.184} & \textcolor{darkgray}{0.231} & 0.467 & 0.524 & 0.341 & 0.483 \\
\midrule
\multicolumn{9}{l}{\textbf{Transformer-based}} \\
Autoformer \cite{wu2021autoformer} & {\cellcolor{gray!15}\textbf{0.056}} & \textcolor{darkgray}{\cellcolor{gray!15}\textbf{0.083}} & \cellcolor{gray!15} \textcolor{darkgray}{\textbf{0.150}} & \cellcolor{gray!15} \textcolor{darkgray}{\textbf{0.193}} & 0.204 & 0.376 & 0.165 & 0.263 \\
FEDformer \cite{zhou2022fedformer} & {\cellcolor{gray!15}\textbf{0.055}} & \textcolor{darkgray}{\cellcolor{gray!15}\textbf{0.081}} & \textcolor{darkgray}{\cellcolor{gray!15}\textbf{0.139}} & \textcolor{darkgray}{\cellcolor{gray!30}\textbf{0.182}} & 0.271 & 0.396 & 0.220 & 0.291 \\ 
Informer \cite{zhou2021informer} & \textcolor{darkgray}{0.070} & \textcolor{darkgray}{0.119} & \textcolor{darkgray}{0.151} & \textcolor{darkgray}{0.199} & 0.283 & 0.324 & 0.137 & 0.217 \\
Reformer \cite{kitaev2020reformer} & \textcolor{darkgray}{0.078} & \textcolor{darkgray}{0.129} & \textcolor{darkgray}{0.234} & \textcolor{darkgray}{0.292} & 0.297 & 0.361 & 0.141 & 0.233 \\
\midrule
\multicolumn{9}{l}{\textbf{RNN/CNN-based}} \\
SFM \cite{zhang2017stock} & \textcolor{darkgray}{0.086} & \textcolor{darkgray}{0.129} & \textcolor{darkgray}{0.161} & \textcolor{darkgray}{0.283} & 0.184 & 0.245 & 0.156 & 0.238 \\
TCN \cite{bai2018empirical} & \textcolor{darkgray}{0.057} & \textcolor{darkgray}{0.083} & \textcolor{darkgray}{0.176} & \textcolor{darkgray}{0.222} & 0.187 & 0.287 & 0.172 & 0.260 \\
LSTNet \cite{lai2018modeling} & \textcolor{darkgray}{0.075} & \textcolor{darkgray}{0.138} & \textcolor{darkgray}{0.148} & \cellcolor{gray!15} \textcolor{darkgray}{\textbf{0.200}} & 0.193 & 0.346 & 0.177 & 0.263 \\
DeepGLO \cite{sen2019think} & {0.090} & \textcolor{darkgray}{0.131} & \textcolor{darkgray}{0.178} & \textcolor{darkgray}{0.400} & 0.264 & 0.372 & 0.181 & 0.246 \\

\midrule
\multicolumn{9}{l}{\textbf{MLP-based}} \\
TimeMixer \cite{wang2024timemixer} & \textcolor{darkgray}{0.091} & \textcolor{darkgray}{0.147} & \textcolor{darkgray}{0.166} & \textcolor{darkgray}{0.211} & 0.181 & 0.314 & 0.167 & 0.279 \\
N-Beats \cite{oreshkin2020nbeats} & \textcolor{darkgray}{0.067} & \textcolor{darkgray}{0.126} & \textcolor{darkgray}{0.162} & \textcolor{darkgray}{0.203} & 0.189 & 0.325 & 0.174 & 0.300 \\
DLinear* \cite{zeng2023dlinear} & \textcolor{darkgray}{0.058} & \textcolor{darkgray}{0.092} & \textcolor{darkgray}{0.257} & \textcolor{darkgray}{0.313} & 0.266 & 0.368 & 0.196 & 0.259 \\
\midrule
\multicolumn{9}{l}{\textbf{Statistic Methods}} \\
VAR \cite{sims1980macroeconomics} & \textcolor{darkgray}{0.096} & \textcolor{darkgray}{0.155} & \textcolor{darkgray}{0.175} & \textcolor{darkgray}{0.222} & 0.243 & 0.381 & 0.177 & 0.260 \\
ARIMA \cite{box1970time} & \textcolor{darkgray}{0.107} & \textcolor{darkgray}{0.169} & \textcolor{darkgray}{0.185} & \textcolor{darkgray}{0.244} & 0.275 & 0.319 & 0.164 & 0.281 \\
S-ARIMA \cite{box2015time} & \textcolor{darkgray}{0.101} & \textcolor{darkgray}{0.163} & \textcolor{darkgray}{0.184} & \textcolor{darkgray}{0.234} & 0.268 & 0.379 & 0.158 & 0.273 \\
\bottomrule
\end{tabular}%
}
\end{table}